\documentclass[11pt]{article}

\usepackage[margin=1in]{geometry}
\usepackage{booktabs}
\usepackage{tabularx}
\newcolumntype{Y}{>{\raggedright\arraybackslash}X}
\usepackage{hyperref}
\usepackage{xcolor}
\usepackage{graphicx}
\usepackage{amsmath}
\usepackage{enumitem}
\usepackage{microtype}
\usepackage[scaled=0.96]{helvet}

\usepackage{setspace}
\usepackage{titlesec}
\usepackage{titling}
\usepackage{tcolorbox}
\tcbuselibrary{skins,breakable}
\usepackage{fancyhdr}
\usepackage{caption}
\usepackage{tikz}
\usepackage{eso-pic}
\usetikzlibrary{arrows.meta,positioning,fit,shapes.geometric,calc}

\definecolor{prismblue}{HTML}{1F5AA6}
\definecolor{deepnavy}{HTML}{0B1020}
\definecolor{electricblue}{HTML}{2563EB}
\definecolor{cyanaccent}{HTML}{06B6D4}
\definecolor{prismgray}{HTML}{4B5563}
\definecolor{mutedgray}{HTML}{6B7280}
\definecolor{lightgray}{HTML}{F3F4F6}
\definecolor{panelbg}{HTML}{F8FAFC}

\setlist{itemsep=0.28em, topsep=0.35em}

\hypersetup{
  colorlinks=true,
  linkcolor=electricblue,
  citecolor=electricblue,
  urlcolor=electricblue,
  pdfauthor={Bin Cao},
  pdftitle={SVGym (SciVerseGym)}
}

\titleformat{name=\section,numberless}
  {\Large\bfseries\color{deepnavy}}
  {}{0pt}{\MakeUppercase}
  [\vspace{-0.45em}{\color{electricblue!70}\rule{0.18\linewidth}{0.8pt}}]
\titlespacing*{\section}{0pt}{1.8em}{0.85em}
\titleformat{name=\subsection,numberless}
  {\large\bfseries\color{electricblue}}
  {}{0pt}{}
\titlespacing*{\subsection}{0pt}{1.25em}{0.45em}

\tcbset{
  sharp corners,
  boxrule=0.45pt,
  left=1.0em,
  right=1.0em,
  top=0.8em,
  bottom=0.8em,
  colback=panelbg,
  colframe=electricblue!32
}

\renewenvironment{abstract}
  {\begin{tcolorbox}[enhanced, colback=panelbg, colframe=electricblue!40, borderline west={1.6pt}{0pt}{electricblue}, title=Abstract, fonttitle=\bfseries\color{deepnavy}, coltitle=deepnavy]\small}
  {\end{tcolorbox}}

\renewcommand{\headrulewidth}{0.25pt}
\renewcommand{\headrule}{\hbox to\headwidth{\color{electricblue!30}\leaders\hrule height \headrulewidth\hfill}}
\title{SVGym (SciVerseGym): An Environment for Reinforcement Learning and Bayesian Optimization in Crystal Discovery}

\author{Bin Cao}

\date{}

\makeatletter
\renewcommand{\maketitle}{%
  \AddToShipoutPictureBG*{%
    \put(\LenToUnit{\paperwidth-2.05cm},\LenToUnit{\paperheight-1.95cm}){%
      \begin{tikzpicture}
        \clip (0,0) circle (0.42cm);
        \node[opacity=0.36, inner sep=0pt] at (0,0)
          {\includegraphics[width=0.84cm,height=0.84cm]{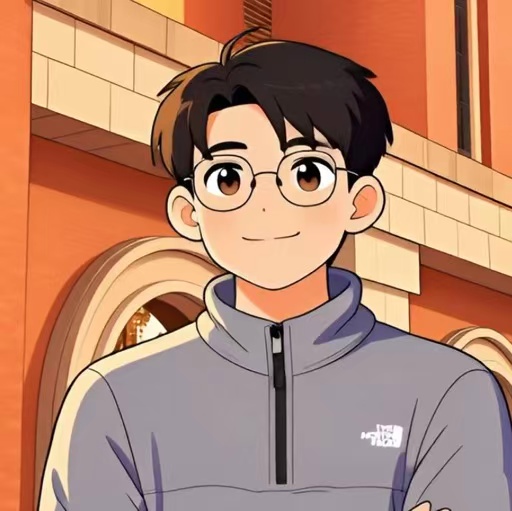}};
        \draw[white, opacity=0.75, line width=0.45pt] (0,0) circle (0.42cm);
      \end{tikzpicture}%
    }%
  }%
  \vspace*{-0.8em}
  \noindent
  \begin{tikzpicture}
    \node[inner sep=0pt, outer sep=0pt, text width=\linewidth] (panel) {\begin{tcolorbox}[
      enhanced,
      colback=white,
      colframe=deepnavy!12,
      boxrule=0.5pt,
      borderline west={2.2pt}{0pt}{electricblue},
      borderline north={0.6pt}{0pt}{cyanaccent!70},
      left=1.25em,right=1.25em,top=1.05em,bottom=1.05em]
      {\footnotesize\bfseries\color{electricblue} SEQUENTIAL MATERIALS DISCOVERY}\par
      \vspace{0.7em}
      {\fontsize{22}{26}\selectfont\bfseries\color{deepnavy}\@title\par}
      \vspace{1.0em}
      {\large\bfseries\color{deepnavy}\@author\par}
      \vspace{0.35em}
      {\small\color{prismgray}The Hong Kong University of Science and Technology (Guangzhou), China \;;\; Huawei Noah's Ark Lab, UK\par}
      {\small\texttt{bcao686@connect.hkust-gz.edu.cn}\par}
      \vspace{0.8em}
      {\scriptsize\color{mutedgray}Gymnasium-Compatible Environments \; \textbullet \; ML Interatomic Potentials \; \textbullet \; BO/RL Agents \; \textbullet \; Crystal Edit Actions}
    \end{tcolorbox}};
  \end{tikzpicture}
  \vspace{0.8em}
}
\makeatother

\begin{document}

\maketitle

\begin{abstract}
Machine-learned interatomic potentials have made atomistic evaluation sufficiently efficient for interactive materials discovery. However, closed-loop crystal search algorithms are still often implemented as isolated pipelines, with bespoke code for structure modification, relaxation, scoring, constraint handling, and experiment bookkeeping. This fragmentation makes it difficult to develop and fairly compare reinforcement learning (RL), Bayesian optimization (BO), evolutionary search, and language-agent workflows under consistent physical assumptions. Here, we introduce SciVerseGym, a Gymnasium-compatible environment for sequential crystal discovery. SciVerseGym formulates crystal design as a Markov decision process in which an agent observes an atomistic structure, applies chemically meaningful edit actions, and receives physically informed feedback from a configurable evaluator. Rather than requiring an agent to generate a complete crystallographic file in a single step, SciVerseGym provides a unified action interface for both local and global structure edits, including elemental substitution, lattice perturbation, atomic displacement, vacancy creation, and atom insertion. The environment supports configurable chemical spaces with broad element sets, user-defined structure pools, atomistic and graph-based observations, customizable reward functions, optional geometry relaxation, and diagnostics such as stability and phonon-related checks. At each step, SciVerseGym applies the selected edit to the current structure, evaluates the resulting candidate using a machine-learned interatomic potential or any ASE-compatible calculator, and returns the standard \texttt{obs, reward, terminated, truncated, info} tuple. By decoupling agent logic from materials-specific infrastructure, SciVerseGym enables diverse search algorithms to be trained, evaluated, and benchmarked within a common framework. SciVerseGym thus provides an open, reproducible, and extensible testbed for rapid prototyping, controlled comparison, and scalable evaluation of closed-loop crystal discovery strategies.

Code is available at: \url{https://github.com/Bin-Cao/SciVerseGym}.

\vspace{0.4em}
\noindent\textbf{Keywords:} Crystal Discovery; Reinforcement Learning; Bayesian Optimization; Gymnasium Environments; SciVerseGym.

\end{abstract}
\newpage
\section*{Introduction}

Computational materials discovery increasingly relies on closed-loop algorithms that can propose a candidate, evaluate it, and use the resulting feedback to choose the next candidate \cite{cao2026physics}. This paradigm is particularly natural for crystalline materials, where a candidate is defined not only by composition, but also by symmetry, lattice parameters, atomic coordinates, defects, and metastable structural motifs. Small changes in any of these variables can alter thermodynamic stability, dynamical stability, synthesizability, and functional properties. Classical high-throughput workflows have been successful because they systematize atomistic calculations and data management. However, they often expose the search loop only indirectly through file generation, workflow engines, relaxation scripts, and post-processing routines. As a result, candidate-generation logic is frequently entangled with implementation-specific details of structure manipulation, calculator setup, constraint handling, and experiment bookkeeping.

At the same time, the algorithmic landscape for materials discovery has broadened. Bayesian optimization is attractive when each evaluation is expensive and sample efficiency is essential \cite{cao2026bgolearn}. Reinforcement learning is attractive when discovery is formulated as a sequence of edits rather than as one-shot generation \cite{yu2026efficient}. Evolutionary algorithms and Monte Carlo methods rely on mutation operators, local moves, and acceptance rules to explore rugged structure--property landscapes. Large language models can suggest chemically meaningful transformations and coordinate scientific workflows, but they are not reliable validators of complete crystallographic files. Although these algorithmic families differ in how they choose candidates, they share a common requirement: a stable interface that accepts a proposed change, applies it to an atomistic structure, evaluates the resulting candidate, and returns machine-readable feedback.

The recent growth of universal and domain-adapted machine-learned interatomic potentials (MLIPs) has made this interface problem more urgent. MLIPs can evaluate energies and forces much faster than density functional theory (DFT), enabling larger candidate sets, repeated relaxation, and more interactive algorithm development \cite{cao2025beyond}. However, MLIP-based discovery introduces backend-specific details, including model checkpoints, supported chemical elements, relaxation settings, reference energies, convex-hull conventions, uncertainty estimates, and optional stability diagnostics such as phonon-related calculations. If every optimization method implements these details separately, results become difficult to reproduce, difficult to compare, and difficult to extend across materials domains.

SciVerseGym addresses this bottleneck by making crystal discovery look like an environment interaction problem while retaining atomistic structures as the central representation. In this formulation, an agent observes the current crystal state, proposes a structured edit, and receives the next state, a scalar reward, termination signals, and diagnostic information. This interaction pattern follows the environment abstraction widely used in reinforcement learning, with Gymnasium providing a standard transition-based API \cite{towers2026gymnasium}. In SciVerseGym, however, the same abstraction is useful beyond reinforcement learning: Bayesian optimization can treat environment calls as structured objective evaluations, evolutionary algorithms can reuse the edit actions as mutation operators, Monte Carlo methods can use them as proposal moves, and language-agent systems can call the environment as a validation and scoring tool rather than generating complete crystal structures in a single step.

SciVerseGym builds on three lines of work. First, large computational materials databases and high-throughput screening platforms have established the value of standardized structure records, calculated properties, and reproducible workflows. Second, MLIPs bridge part of the cost gap between DFT and empirical potentials, making repeated relaxation and scoring feasible inside algorithmic loops. Third, environment APIs from reinforcement learning provide a clean separation between decision-making logic and task-specific transition dynamics. SciVerseGym combines these ideas into a materials-specific environment layer for \textbf{closed-loop crystal discovery.}

The contribution of SciVerseGym is not a new force field, a new crystal generator, or a new optimizer. Its contribution is an integration layer for materials search: a bounded action schema for chemically meaningful structure edits, a Gymnasium-compatible transition protocol, calculator adapters for atomistic evaluation, observation builders for atomistic and graph-based states, configurable reward functions, optional relaxation and diagnostic checks, and baseline examples that allow different search methods to operate on the same atomistic task. By decoupling search algorithms from materials-specific infrastructure, SciVerseGym provides a common language for comparing reinforcement learning, Bayesian optimization, evolutionary search, Monte Carlo exploration, and language-agent strategies under consistent physical assumptions.

\section*{Design principles}

SciVerseGym is guided by four design principles.

\begin{enumerate}[leftmargin=*]
\item \textbf{Structured edits instead of raw file generation.}
Agents interact with crystals through bounded and interpretable edit actions
rather than by emitting complete crystallographic files. These actions encode
common structure modifications, such as elemental substitution, lattice
perturbation, atomic displacement, vacancy creation and atom insertion. This
design reduces invalid outputs, makes the search space easier to constrain,
and allows the same interface to be used by reinforcement-learning policies,
Bayesian optimizers, evolutionary algorithms, language agents and human users.

\item \textbf{A standard closed-loop transition interface.}
SciVerseGym follows the Gymnasium interaction protocol, in which each action
produces an observation, reward, termination flag, truncation flag and
auxiliary information. This standard transition format allows existing
reinforcement-learning and optimization tools to interact with the materials
discovery task without method-specific wrappers, while preserving the
sequential nature of crystal editing.

\item \textbf{Separation between search policy and physical evaluation.}
The search algorithm is decoupled from the calculator used to evaluate a
candidate structure. The same agent can therefore operate with different
backends, including SevenNet \cite{park_scalable_2024}, MatterSim\cite{yang2024mattersim}, ORB \cite{neumann2024orb}, surrogate models, reference
calculators or custom ASE-compatible calculators. Backend choices, relaxation
settings, reference data and optional diagnostics are specified when the
environment is constructed, rather than being hard-coded into the search
method.

\item \textbf{Transparent and extensible feedback.}
The scalar reward is returned together with physically meaningful diagnostic
information whenever the selected evaluator supports it. This information can
include total energy, formation energy, energy above the convex hull, phonon
stability, relaxation status and backend metadata. Providing these quantities
through the \texttt{info} dictionary makes optimization behavior easier to
interpret, debug and compare across different algorithms and evaluator
configurations.
\end{enumerate}

\section*{Framework overview}

\begin{figure}[htbp]
    \centering
    \includegraphics[width=1\columnwidth]{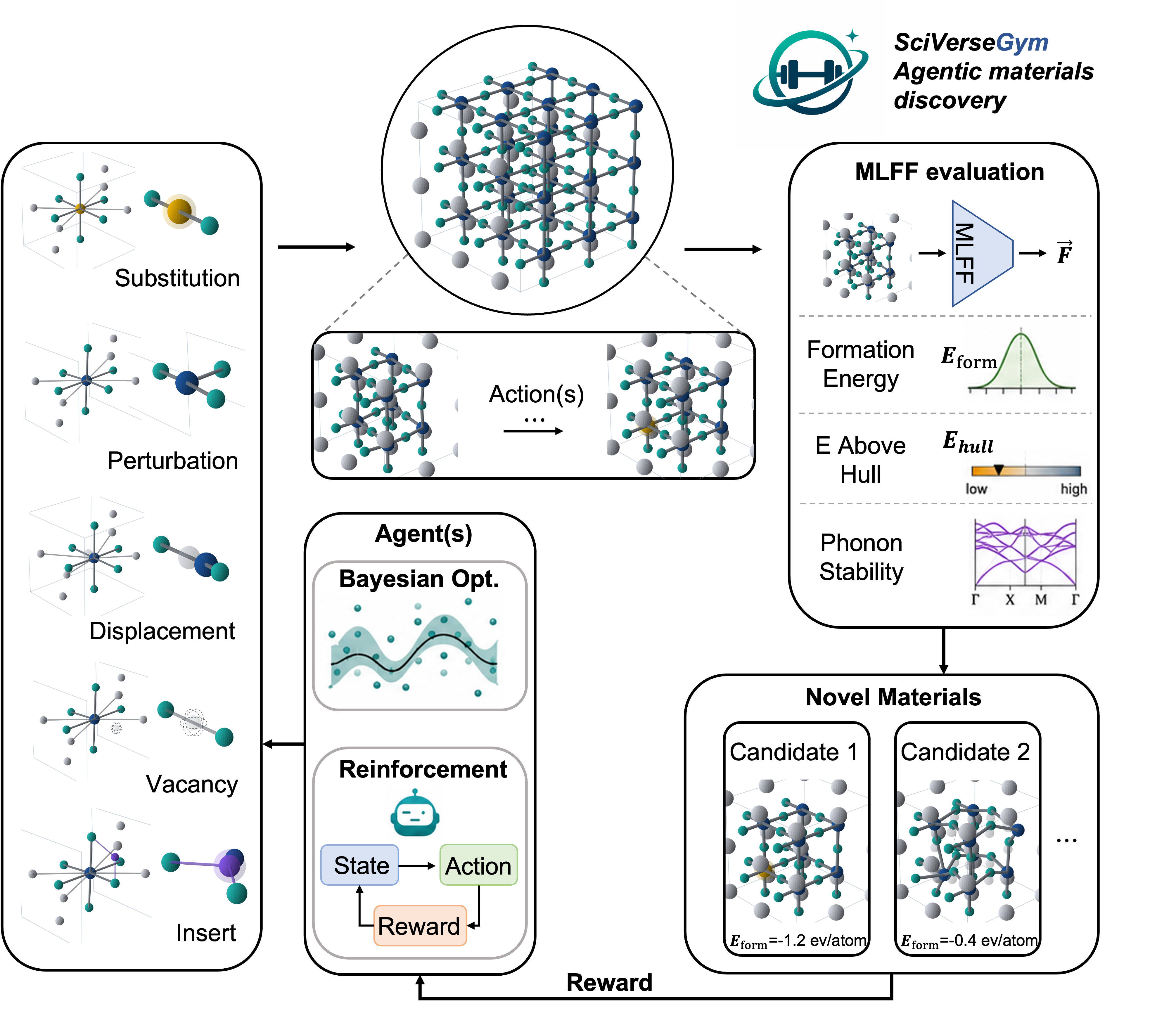}
    \caption{\textbf{Closed-loop structure of SciVerseGym.} A search algorithm
    proposes a structured crystal edit rather than a complete crystal file. The
    environment applies the edit to the current atomistic state, calls the
    selected evaluator, optionally uses reference data for derived physical
    quantities, and returns a standardized transition. The same interface can be
    used by Bayesian optimization, reinforcement learning, evolutionary search,
    Monte Carlo methods, and language-agent workflows.}

    \label{fig:ATOMmodel}
\end{figure}

SciVerseGym exposes crystal discovery as a sequential decision process over
atomistic structures (Figure \ref{fig:ATOMmodel}). Instead of treating materials design as a one-shot
generation problem, the environment represents discovery as a finite sequence
of chemically meaningful edits applied to an initial crystal. At step ($t$), an
agent observes the current state ($s_t$), selects a structured action ($a_t$),
and receives a standard environment transition,

\begin{equation}
s_{t+1}, r_t, d_t, \tau_t, i_t =
\mathrm{Env.step}(a_t),
\end{equation}

where $s_{t+1}$ is the next observation, $r_t$ is the scalar reward,
$d_t$ is the termination flag, $\tau_t$ is the truncation flag, and $i_t$
is a dictionary of diagnostic information. This interface follows the
Gymnasium convention while specializing the state, action, reward, and
diagnostic fields for crystal discovery. The current implementation registers
the environment as \texttt{CrystalDiscovery-v0}.

An episode begins by sampling or selecting a structure from a dataset. The
default workflow uses structures from ALEX-MP-20 \cite{hollmer2025open}, but user-defined datasets can
also be mounted if they provide atomistic structures and associated metadata.
The environment stores the current crystal as an ASE \texttt{Atoms} object and
uses it as the central state representation throughout the episode. At every
step, the selected action is first interpreted as a bounded crystal edit. The
edit is applied to the current structure to produce a candidate structure. The
candidate is then evaluated by the configured backend, which may be a
machine-learned interatomic potential, a reference-backed physical calculator,
a surrogate model, or any ASE-compatible calculator. The resulting energy,
optional relaxed structure, reward, and diagnostic metadata are returned to the
agent.

The observation ($s_t$) is designed to support both atomistic and learning-based
agents. It can include the current ASE structure, graph-based representations
when graph construction is available, scalar evaluation results such as total
energy, and bookkeeping variables such as the current step index and remaining
edit budget. The \texttt{info} dictionary provides additional physical and
backend-specific information that should not necessarily be compressed into the
scalar reward. Depending on the selected evaluator and reference data, this may
include the chemical formula, total energy, formation energy per atom, energy
above the convex hull, phonon-stability indicators, relaxation status, backend
name, and other metadata useful for debugging and analysis.

The distinction between \texttt{terminated} and \texttt{truncated} follows the
standard environment semantics. A transition may be terminated when the task
defines a natural stopping condition, for example when a target stability or
property criterion is reached. A transition is truncated when the episode stops
because of an external limit, most commonly the maximum number of allowed edit
steps. This separation is important for reinforcement-learning algorithms, but
it is also useful for Bayesian optimization, evolutionary search, and language
agents because it records whether a trajectory ended because it succeeded,
failed, or simply exhausted its search budget.

This design separates three responsibilities that are often entangled in
materials-discovery code. The search agent is responsible for choosing actions.
The environment is responsible for applying actions, maintaining episode state,
checking basic validity constraints, and packaging observations. The evaluator
is responsible for assigning physical quantities such as energy, forces,
formation energy, hull distance, or phonon stability when the corresponding
backend supports them. Because these components are separated, the same search
policy can be tested with different MLIP backends, different reward functions,
different datasets, or different reference databases without rewriting the
agent itself.

\subsection*{Action space}

The action space is designed to modify crystal composition and geometry without requiring agents to generate complete crystallographic files. Each action is represented as a compact dictionary containing an \texttt{action\_type} field and a small number of typed arguments. The \texttt{action\_type} specifies the edit operator, while the remaining fields define the target site, chemical element, displacement vector, lattice perturbation, or insertion coordinate. This representation is intentionally simple and algorithm-agnostic: it can be sampled by reinforcement-learning policies, enumerated by Bayesian-optimization baselines, mutated by evolutionary algorithms, proposed by language agents, or inspected directly by human researchers.

The current implementation supports five action families. Element replacement selects an existing atomic site and changes its chemical species, making it suitable for substitutional search, dopant discovery, and local exploration of composition space. Lattice perturbation modifies the unit cell through a bounded change to the lattice degrees of freedom, enabling exploration of strain, volume, and cell-shape effects without reconstructing the full structure. Atomic displacement moves a selected atom by a bounded vector and is useful for exploring local distortions, off-center motifs, and nearby metastable configurations. Vacancy creation removes an atom from a selected site, allowing the search process to explore defects and stoichiometric changes. Atom insertion adds a selected element at a fractional coordinate, supporting the exploration of interstitial sites and composition growth.

\begin{itemize}[leftmargin=*]
\item \textbf{Element replacement.}
This action replaces the chemical species at an existing atomic site and is
mainly used for substitution, alloying, doping, and local compositional
exploration. 

\item \textbf{Lattice perturbation.}
This action applies a bounded perturbation to the unit cell, enabling
exploration of strain-like moves, volume changes, and cell-shape variations.

\item \textbf{Atomic displacement.}
This action moves a selected atom by a bounded displacement vector. It
supports local structural distortion, motif perturbation, and
relaxation-like exploration around the current configuration. 

\item \textbf{Vacancy creation.}
This action removes an atom from a selected site, allowing the search process
to explore point defects, stoichiometric changes, and the removal of
unfavorable atoms. 

\item \textbf{Atom insertion.}
This action inserts a selected element at a fractional coordinate, supporting
composition growth, interstitial-site exploration, and site discovery. 
\end{itemize}

Validity handling is implemented at the environment level. Atomic sites are mapped to valid indices of the current structure. Element identifiers are interpreted as atomic numbers and clipped to the supported range. Lattice perturbations and atomic displacements are bounded to prevent early-stage agents from producing extreme or numerically unstable structures. Inserted atoms are specified in fractional coordinates, making the action independent of the absolute cell scale. Vacancy actions preserve at least one atom so that the environment does not collapse into an empty structure. These rules do not guarantee that every generated candidate is physically realistic, but they reduce common syntactic and numerical failures and prevent invalid actions from dominating algorithm development.

This bounded action interface is important for fair comparison across search methods. If different algorithms use different hand-written mutation operators or different validity rules, performance differences may reflect implementation details rather than the search strategy itself. In SciVerseGym, reinforcement-learning agents, Bayesian optimizers, evolutionary methods, Monte Carlo samplers, and language-agent workflows can operate through the same edit operators and receive feedback from the same evaluator. Algorithms may therefore differ in how they choose actions, while the structure edits, validity handling, evaluation protocol, and returned diagnostics remain controlled.

\subsection*{Observation and feedback}

At each environment step, SciVerseGym returns a standard transition containing an observation, scalar reward, termination flag, truncation flag, and diagnostic information. The observation contains the current atomistic state and the information needed by an agent to select the next action. In the default implementation, this includes the current ASE \texttt{Atoms} object, an optional graph representation, the current energy, the step count, the remaining edit budget, and metadata associated with the structure or dataset entry.

The \texttt{info} dictionary is deliberately richer than the scalar reward. It records quantities that are useful for physical interpretation, debugging, benchmarking, and post-processing, even when they are not directly used by the default reward. Depending on the selected evaluator and available reference data, \texttt{info} can include the backend identity, chemical formula, total energy, formation energy per atom, energy above the convex hull, phonon-stability indicators, minimum phonon frequency, relaxation status, and backend-specific metadata.

This separation between observation, reward, and diagnostic information is central to the design of SciVerseGym. Learning algorithms can optimize a scalar reward, while researchers can inspect the diagnostic fields to understand why a candidate performed well or poorly. The same transition can also be reused to define alternative objectives. For example, one study may optimize energy above hull, another may prioritize novelty or elemental abundance, and a later screening stage may incorporate phonon stability or elastic properties, all without changing the high-level environment interface.

\subsection*{Evaluation backends}

SciVerseGym separates the environment loop from the physical evaluator. The recommended workflow selects a machine-learned force-field backend during environment construction. Built-in adapters are provided for SevenNet \cite{park_scalable_2024}, MatterSim\cite{yang2024mattersim} and ORB \cite{neumann2024orb} when the corresponding packages and checkpoints are available. Users may also provide a custom ASE-compatible calculator, which allows the same environment interface to be used with surrogate models, reference-backed calculators, empirical potentials, or higher-fidelity atomistic methods. If no physical backend is supplied, a deterministic surrogate calculator can be used for smoke tests and interface development, but it should not be interpreted as a scientific energy model.

A force field directly provides quantities such as total energy and forces. Derived thermodynamic quantities require additional reference data. Formation energy requires same-force-field elemental reference energies, and convex-hull distance requires compatible reference phases evaluated under the same backend and convention. When these references are unavailable, SciVerseGym returns \texttt{None} rather than fabricating thermodynamic quantities. This behavior makes missing information explicit and prevents downstream algorithms from silently optimizing inconsistent physical targets.

For a candidate structure with total energy \(E\), element counts \(n_i\), elemental reference energies \(\mu_i\), and total atom count \(N\), the formation energy per atom is computed as

\begin{equation}
E_{\mathrm{form}} =
\frac{E - \sum_i n_i\mu_i}{N}.
\end{equation}

Energy above hull is computed by comparing this formation energy with the lowest compatible combination of same-force-field reference phases at the same composition. Phonon stability, when enabled, is estimated using finite-displacement force calculations through Phonopy \cite{togo2023first}. These optional diagnostics allow SciVerseGym to support both fast algorithm development and more physically informed screening without requiring changes to the agent logic.

\subsection*{Reward design}

The current default reward is deliberately transparent and lightweight:

\begin{equation}
r =
0.80\,r_{\mathrm{stability}}
+ 0.10\,r_{\mathrm{abundance}}
+ 0.10\,r_{\mathrm{novelty}}.
\end{equation}

The stability term is derived from same-force-field energy above hull when that quantity is available; otherwise, it is set to zero. The abundance term is a simple proxy based on average atomic number, encouraging candidates composed of lighter and more abundant elements. The novelty term rewards chemical formulas that have not yet appeared in the current episode, encouraging the search process to explore new regions of composition space rather than repeatedly revisiting the same formula.

Phonon stability is not included in the current default reward. It may still be reported by the evaluator through the \texttt{info} dictionary and can be incorporated into custom objectives in future studies. This distinction is intentional: the default reward provides a transparent reference objective for benchmarking and development, while the environment interface remains flexible enough to support task-specific reward functions.

The reward should therefore be viewed as a reference objective rather than a universal materials-design target. Different materials-discovery tasks may require different trade-offs between stability, novelty, synthesizability, cost, abundance, mechanical properties, or functional performance. SciVerseGym is designed so that users can replace the reward function without changing the action schema, observation format, evaluator interface, or closed-loop transition protocol.

\section*{Agent protocol}

SciVerseGym defines a common interaction protocol for search agents. An agent
does not need to know how structures are stored, how calculators are executed,
or how diagnostic quantities are computed. Instead, it observes the current
state, selects a structured crystal-edit action, submits the action to the
environment, and updates its internal model from the returned transition. At
step \(t\), the agent receives an observation \(s_t\), chooses an action
\(a_t \in \mathcal{A}(s_t)\), and obtains $s_{t+1}, r_t, d_t, \tau_t, i_t = \mathrm{Env.step}(a_t)$,
where \(r_t\) is the scalar reward, \(d_t\) is the termination flag,
\(\tau_t\) is the truncation flag, and \(i_t\) contains diagnostic information
such as backend identity, formula, energy, formation energy, energy above hull,
or phonon-related quantities when available. This protocol separates the search
logic from materials-specific infrastructure. Different agents may use
different policies, surrogate models, acquisition functions, or memory
mechanisms, but they interact with the same action schema, evaluator, reward
definition, and transition format.

\subsection*{Bayesian-optimization agent}

SciVerseGym can be used with Bayesian optimization by treating each environment
step as a structured black-box evaluation. In this setting, the BO agent first
constructs a discrete candidate set of valid edits
\(\mathcal{A}_t = \{a_t^{(1)}, \ldots, a_t^{(M)}\}\) from the current
observation. Each state-action pair is then mapped to a fixed-dimensional
feature vector,

\begin{equation}
  x_t^{(j)} = \phi(s_t, a_t^{(j)}),
\end{equation}

where \(\phi\) may include composition descriptors, atomic numbers, action type,
target-site information, energy history, graph features, or handcrafted
structure features. A surrogate model is fitted to the accumulated data and
used to predict the expected reward and uncertainty of each candidate action.
For example, an upper-confidence-bound acquisition function can be written as

\begin{equation}
  \alpha_t(a) =
  \mu_t\!\left(\phi(s_t, a)\right)
  + \kappa_t\sigma_t\!\left(\phi(s_t, a)\right),
\end{equation}

where \(\mu_t\) and \(\sigma_t\) are the surrogate posterior mean and
uncertainty, and \(\kappa_t\) controls the exploration--exploitation trade-off \cite{frazier2018tutorial}. The agent selects

\begin{equation}
  a_t^\star =
  \arg\max_{a \in \mathcal{A}_t} \alpha_t(a),
\end{equation}

submits \(a_t^\star\) to the environment, and records the returned reward and
next state. The BO history is updated as

\begin{equation}
  \mathcal{D}_{t+1} =
  \mathcal{D}_t \cup
  \{(s_t, a_t^\star, r_t, s_{t+1}, i_t)\}.
\end{equation}

This usage is particularly useful when the evaluator is expensive, the number
of allowed crystal edits is limited, or sample efficiency is more important
than long-horizon policy learning. Because the action space is structured, the
BO agent can enumerate or sample chemically meaningful edits instead of
optimizing over unconstrained crystal files. The same protocol also makes it
straightforward to compare different feature maps, surrogate models, and
acquisition functions under identical environment settings.

\subsection*{Reinforcement-learning agent}

Reinforcement learning treats crystal discovery as a sequential editing
problem. A policy observes the current state, selects a crystal-edit action,
receives a reward, and continues until the episode terminates or the edit
budget is exhausted. The policy may be deterministic or stochastic,

\begin{equation}
  a_t \sim \pi_\theta(a \mid s_t),
\end{equation}

where \(\theta\) denotes the learnable parameters. The resulting trajectory is

\begin{equation}
  \tau = (s_0, a_0, r_0, s_1, a_1, r_1, \ldots),
\end{equation}

and the agent can be trained to maximize the discounted return

\begin{equation}
  G_t = \sum_{k=0}^{T-t} \gamma^k r_{t+k},
\end{equation}

where \(\gamma\) is the discount factor and \(T\) is the episode length. The
same environment interface can support simple tabular or action-value methods,
policy-gradient algorithms, actor--critic methods, graph neural-network
policies, or hybrid language-model policies. These methods differ in how they
represent \(s_t\) and how they choose \(a_t\), but they all consume the same
standardized transition

\begin{equation}
  (s_t, a_t, r_t, s_{t+1}, d_t, \tau_t, i_t).
\end{equation}

An RL agent may learn from graph embeddings, handcrafted composition and
structure descriptors, energy histories, diagnostic metadata, or multimodal
state summaries. For example, a graph neural-network policy can use atomic
numbers, local environments, and connectivity as input, whereas a simpler
baseline can use action values over a controlled candidate-action set. A
language-model-assisted policy can propose high-level chemical edits, but the
environment remains responsible for applying the edit, validating the action,
evaluating the candidate, and returning numerical feedback.

The bounded action schema is important for RL stability. Since actions
correspond to interpretable structure edits, invalid operations are easier to
clip, reject, or diagnose than in free-form crystal generation. This reduces
the burden on the policy during early training and allows failures to be
recorded through the \texttt{info} dictionary rather than hidden inside
task-specific scripts. As a result, RL methods can focus on learning search
strategies while SciVerseGym handles structure bookkeeping, evaluation, reward
calculation, and transition formatting.

\section*{Implementation and availability}

SciVerseGym is implemented in Python and distributed as an open-source repository. The package is built around the Gymnasium environment protocol and uses ASE for atomistic structure representation and calculator interoperability. It also relies on standard scientific Python libraries for numerical computation and tabular data handling, with optional support for graph-style observations, convex-hull analysis, and phonon-related diagnostics through external packages such as PyTorch Geometric, SciPy, and Phonopy. When the package is imported, it registers the \texttt{CrystalDiscovery-v0} environment, which provides the main interface for closed-loop crystal discovery.

The implementation is designed to keep the agent interface stable while allowing backend details to evolve. Users can configure the environment with machine-learned interatomic potentials, reference-backed physical calculators, surrogate evaluators, or custom ASE-compatible calculators. Optional components such as geometry relaxation, convex-hull references, phonon diagnostics, graph observations, and reward definitions can be enabled or modified depending on the scientific task. This modular structure allows new evaluators, datasets, action constraints, and diagnostic quantities to be added without changing the conceptual environment protocol.

The repository (\url{https://github.com/Bin-Cao/SciVerseGym}) includes source code, tests, example Bayesian-optimization and reinforcement-learning baselines, dataset utilities, calculator adapters, and documentation. Detailed installation instructions, environment configuration options, action schemas, reward definitions, formation-energy and convex-hull conventions, phonon settings, dataset formats, and troubleshooting notes are provided in the project documentation.

\section*{Limitations}

SciVerseGym is an interface and benchmarking substrate, not a guarantee of
physical validity. Its predictions are only as reliable as the selected
calculator, reference data and validation protocol. Energy above hull is
meaningful only when candidate and reference phases are evaluated in a
consistent energy convention. Phonon stability is optional and can be
computationally expensive. The default reward is intentionally simple and should
be replaced for domain-specific studies. Finally, structured edit actions make
the search problem easier to validate, but they do not enforce charge balance,
synthesis feasibility, or all forms of chemical
reasonableness.

\section*{Discussion}

The central claim of SciVerseGym is that materials-search algorithms should be
compared through a common interaction protocol. Such a protocol does not remove
the need for domain expertise, careful reference data or high-fidelity
validation. It does, however, make algorithmic experiments easier to reproduce:
the same starting structures, action schema, evaluator and reward can be shared
across Bayesian optimization, reinforcement learning, evolutionary search and
language-agent workflows. This separation should help materials researchers
focus on scientific objectives while allowing algorithm developers to test
policies without rewriting atomistic infrastructure.

\section*{Acknowledgements}

SciVerseGym builds on the open scientific Python ecosystem, including
Gymnasium, ASE, NumPy, SciPy, PyTorch Geometric and Phonopy, as well as modern
machine-learned force-field packages.

\bibliographystyle{unsrt}
\bibliography{references}

\end{document}